\documentclass[conference]{IEEEtran}
\usepackage{amsmath,amssymb,amsfonts}
\usepackage{algorithmic}
\usepackage{graphicx, subfigure}
\usepackage{textcomp}
\usepackage{xcolor}
\usepackage[hyphens]{url}
\def\BibTeX{{\rm B\kern-.05em{\sc i\kern-.025em b}\kern-.08em
    T\kern-.1667em\lower.7ex\hbox{E}\kern-.125emX}}
\usepackage{hyperref}
\usepackage[capitalise]{cleveref}
\usepackage[style=ieee,backend=biber,maxnames=6]{biblatex}
\begin{document}

\title{Turning LLM Activations Quantization-Friendly}

\author{%
  \begin{minipage}[t]{0.32\linewidth}
    \centering
    Patrik Czakó\\[0.5ex]
    \textit{Doctoral School of Applied Informatics and Applied Mathematics, Obuda University}\\
    Budapest, Hungary\\
    czako.patrik@stud.uni-obuda.hu
  \end{minipage}\hfill
  \begin{minipage}[t]{0.32\linewidth}
    \centering
    Gábor Kertész\\[0.5ex]
    \textit{John von Neumann Faculty of Informatics, Obuda University}\\
    Budapest, Hungary\\
    kertesz.gabor@nik.uni-obuda.hu
  \end{minipage}\hfill
  \begin{minipage}[t]{0.32\linewidth}
    \centering
    Sándor Szénási\\[0.5ex]
    \textit{John von Neumann Faculty of Informatics, Obuda University}\\
    Budapest, Hungary\\
    szenasi.sandor@nik.uni-obuda.hu
  \end{minipage}%
}

\maketitle

\begin{abstract}
Quantization effectively reduces the serving costs of Large Language Models (LLMs) by speeding up data movement through compressed parameters and enabling faster operations via integer arithmetic. However, activating integer arithmetic requires quantizing both weights and activations, which poses challenges due to the significant outliers in LLMs that increase quantization error. In this work, we investigate these outliers with an emphasis on their effect on layer-wise quantization error, then examine how smoothing and rotation transform the observed values. Our primary contributions include introducing a new metric to measure and visualize quantization difficulty based on channel magnitudes, as well as proposing a hybrid approach that applies channel-wise scaling before rotation, supported by a mathematical formulation of its benefits. The code used in our experiments is available at \url{https://github.com/czakop/smooth-rot-exper}.

\end{abstract}


\section{Introduction}

Large Language Models (LLMs) continue to evolve, with open-source models such as LLaMA2~\cite{touvronLlama2Open2023} and LLaMA3~\cite{grattafioriLlama3Herd2024} emerging. However, in contrast to smaller predecessors \cite{yangTrainingExperimentalLanguage2023, voFakeNewsDetection2025}, these billions-parameter LLMs create substantial deployment challenges on resource-constrained devices, particularly regarding memory usage and inference speed. Quantization is a widely used technique to compress neural networks \cite{wangModelCompressionEfficient2024, zhuSurveyModelCompression2024}, resulting in smaller model sizes and accelerated inference. To take full advantage of these benefits, we need to quantize both weights (the model’s learned parameters) and activations (the intermediate layer outputs), enabling integer matrix multiplications \cite{jacobQuantizationTrainingNeural2017}. Nonetheless, LLMs exhibit activation outliers \cite{dettmersLLMint88bitMatrix2022, xiaoSmoothQuantAccurateEfficient2023, yangMitigatingQuantizationErrors2024, linDuQuantDistributingOutliers2024}, which substantially increase quantization error.

Researchers have explored various strategies to address this challenge \cite{dettmersLLMint88bitMatrix2022, zhaoAtomLowbitQuantization2024, xiaoSmoothQuantAccurateEfficient2023, ashkboosQuaRotOutlierFree4Bit2024}, and equivalent transformations appear especially promising. Among them, channel-wise scaling~\cite{xiaoSmoothQuantAccurateEfficient2023, shaoOMNIQUANTOMNIDIRECTIONALLYCALIBRATED2024} and rotation~\cite{ashkboosQuaRotOutlierFree4Bit2024, linDuQuantDistributingOutliers2024, liuSpinQuantLLMQuantization2024} are the most frequently adopted. Despite their practical effectiveness, a notable performance gap remains between these approaches and weight-only quantization techniques~\cite{dettmersSPQRSPARSEQUANTIZEDREPRESENTATION2024, kimSqueezeLLMDenseandSparseQuantization2024, tsengQuIPEvenBetter2024, chenEfficientQATEfficientQuantizationAware2024}, where activations remain unquantized. This paper seeks to help bridge that gap by investigating how activation outliers influence quantization error, as well as examining the impact of these transformations on outliers, and thus on overall quantization performance. In summary, our main contributions are as follows:

\begin{itemize}
    \item We introduce a new metric to measure and visualize quantization difficulty based on channel magnitudes.
    \item We highlight that although rotation typically yields lower quantization error than channel-wise scaling, it can underperform even the untransformed model when massive outliers are present.
    \item We propose applying channel-wise scaling before rotation to achieve the best of both worlds, backed by a mathematical formulation that clarifies its effects on outliers.
\end{itemize}

\section{Preliminaries}

\subsection{Symmetric Quantization}

Quantization reduces the number of bits used to represent activations and weights in neural networks. The process begins by determining a quantization grid for the floating-point tensor, consisting of a smaller set of discrete values that the quantized tensor can take. In this work, we employ symmetric integer quantization, which provides a uniform grid symmetric around zero. The next step involves mapping the original floating-point numbers to the quantization grid. This is typically achieved using the Round-To-Nearest (RTN) method, where the normalized values are rounded to the nearest grid points. Alternatively, advanced techniques such as GPTQ~\cite{frantarOPTQACCURATEPOSTTRAINING2023} leverage gradient information to optimize this step. The symmetric integer quantization process, when using $b$ bits per value and the RTN method, can be expressed as follows:

\begin{equation} \label{eq:sym_quant}
    \mathbf{X}_{INT} = \lfloor \frac{\mathbf{X}}{\Delta} \rceil, \quad \Delta = \frac{\max(\lvert\mathbf{X}\rvert)}{2^{b-1} -1}.
\end{equation}

Here, $\mathbf{X}$ is the real-valued input tensor, $\mathbf{X}_{INT}$ is the corresponding point on the integer grid, $\Delta$ is the quantization step size, and $\lfloor \cdot \rceil$ represents the rounding function. The quantized value is then obtained by scaling back: $Q(\mathbf{X}) = \mathbf{X}_{INT} \cdot \Delta$. This process enables efficient computation while maintaining a close approximation of the original values.

\subsection{Layer-wise Quantization Error}

The layer-wise quantization error is defined as the squared Frobenius norm of the difference between the original and quantized layer outputs:

\begin{equation} \label{eq:quant_error}
 Error_Q\!\left(\mathbf{X},\mathbf{W}\right) = \left\| \mathbf{X} \mathbf{W} - Q\!\left(\mathbf{X}\right) Q\!\left(\mathbf{W}\right) \right\|_F^2,
\end{equation}

where $\mathbf{X} \in \mathbb{R}^{n \times c_{in}}$ is the original activation tensor with sequence length $n$ and embedding dimensions $c_{in}$, and $\mathbf{W} \in \mathbb{R}^{c_{in} \times c_{out}}$ is the weight matrix mapping $\mathbf{X}$ into $c_{out}$ output channels. This error is proportional to the quantization noise of the activations, amplified by the norm of the weight matrix, and vice versa. Assuming the quantization noise is uniformly distributed within $[-\frac{\Delta}{2}, \frac{\Delta}{2}]$, its variance is given by $\frac{\Delta^2}{12}$. Consequently, the layer-wise quantization error can be reduced by lowering the norms of the matrices and/or decreasing the quantization step size. Since the step size is determined by the maximum value of the tensor (as shown in \eqref{eq:sym_quant}), tensors with smoother distributions and fewer outliers result in lower quantization errors.

FlatQuant~\cite{sunFlatQuantFlatnessMatters2024} argues that tensors with lower kurtosis lead to reduced quantization error. To quantify flatness, FlatQuant visualizes sorted channel magnitudes (Frobenius norm) of weights and activations seeking flatter distributions. Building on this approach, we measure quantization difficulty by calculating the standard deviation of the channel magnitudes.

\subsection{Equivalent Transformations}

Input activations of different linear modules in large language models exhibit significant outliers~\cite{dettmersLLMint88bitMatrix2022, xiaoSmoothQuantAccurateEfficient2023}, which pose challenges for quantization as described above. To address these difficulties, researchers have proposed mitigating activation outliers by applying affine transformations to the input tensor. In order to ensure numerical equivalence, the weight tensor must also undergo an inverse transformation:

\begin{equation} \label{eq:eq_transformation}
\mathbf{Y} = \mathbf{X} \mathbf{W} = \mathbf{X}\underbrace{\left( \mathbf{A}\mathbf{A}^{-1}\right)}_{\mathbb{I}} \mathbf{W} = \underbrace{\left(\mathbf{X} \mathbf{A}\right)}_{\hat{\mathbf{X}}} \cdot \underbrace{\left(\mathbf{A}^{-1} \mathbf{W}\right)}_{\hat{\mathbf{W}}},
\end{equation}

where $\mathbf{A} \in \mathbb{R}^{c_{in} \times c_{in}}$ denotes the invertible matrix of the linear transformation, $\mathbb{I}$ represents the identity matrix of size $c_{in}$, and $\hat{\mathbf{X}}$  and $\hat{\mathbf{W}}$ are the transformed input and weight matrices, respectively. The primary objective of these techniques is to design $\mathbf{A}$ in a way that minimizes quantization error $Error_Q(\hat{\mathbf{X}},\hat{\mathbf{W}})$. While certain methods focus on directly optimizing $\mathbf{A}$~\cite{sunFlatQuantFlatnessMatters2024}, two widely adopted approaches simplify the design process to facilitate efficient construction of $\mathbf{A}$. One approach is channel-wise scaling (also known as smoothing) ~\cite{xiaoSmoothQuantAccurateEfficient2023, shaoOMNIQUANTOMNIDIRECTIONALLYCALIBRATED2024}, where $\mathbf{A}$ is artificially made diagonal and defined by the scaling factor $\mathbf{s} \in \mathbb{R}^{c_{in}}$, such that $\mathbf{A}^{-1} = \mathrm{diag}(\mathbf{s})$. In this method, each channel of $\mathbf{X}$ is divided by a constant before quantization, while the corresponding channels of $\mathbf{W}$ are multiplied by the same factor. The other main technique is rotation~\cite{ashkboosQuaRotOutlierFree4Bit2024, linDuQuantDistributingOutliers2024, liuSpinQuantLLMQuantization2024}, where an orthogonal matrix $\mathbf{R}$ is used for the transformation, satisfying $\mathbf{R}\mathbf{R}^\intercal = \mathbb{I}$, so $\mathbf{A}^{-1} = \mathbf{R}^\intercal$. Unlike smoothing, rotation redistributes outlier values across channels.
\section{Experimental Setup}

\begin{figure*}[t]
    \centering
    \subfigure[Original]{
        \includegraphics[width=0.225\linewidth]{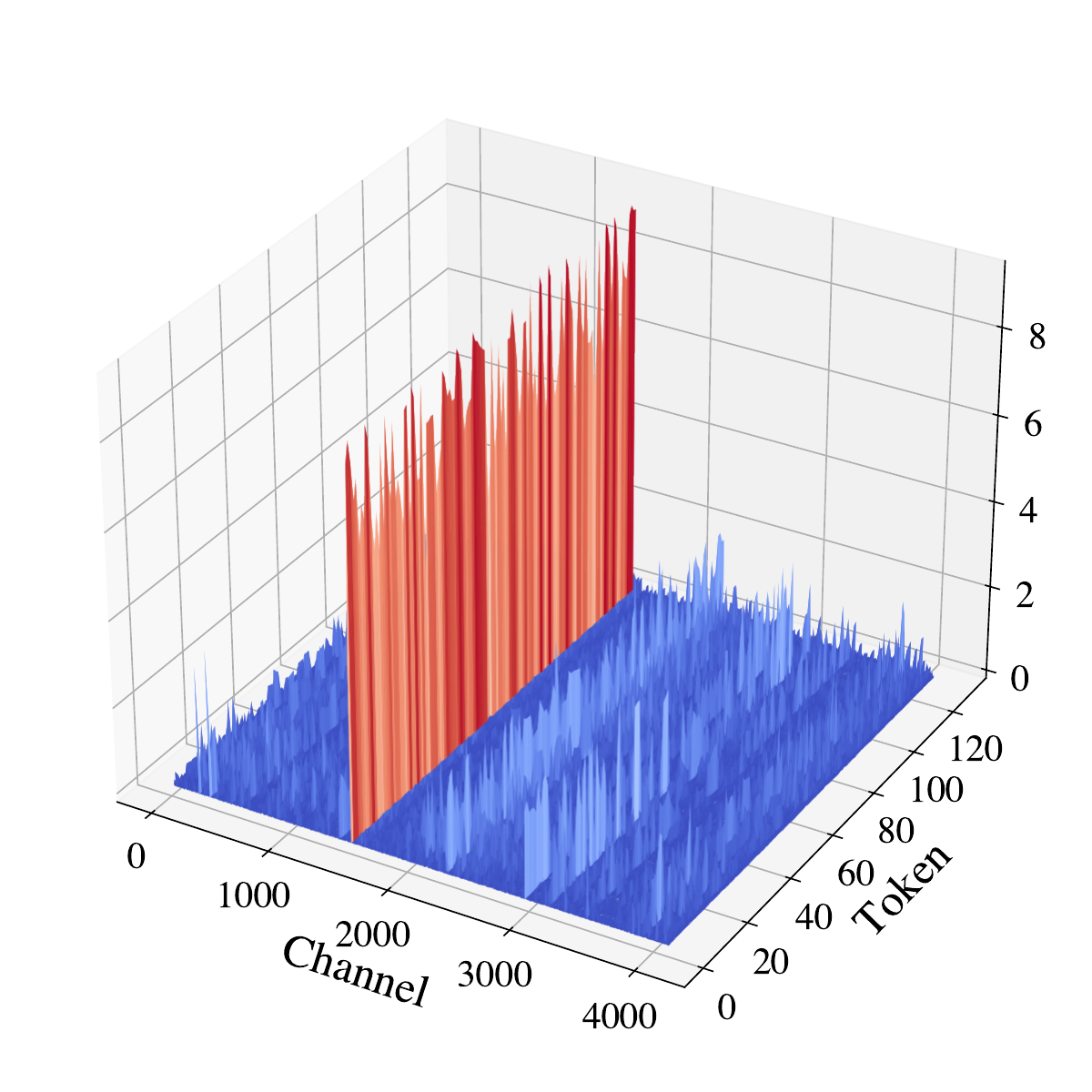}
        \label{fig:act_k_1_orig}
    }
    \subfigure[Smoothing]{
        \includegraphics[width=0.225\linewidth]{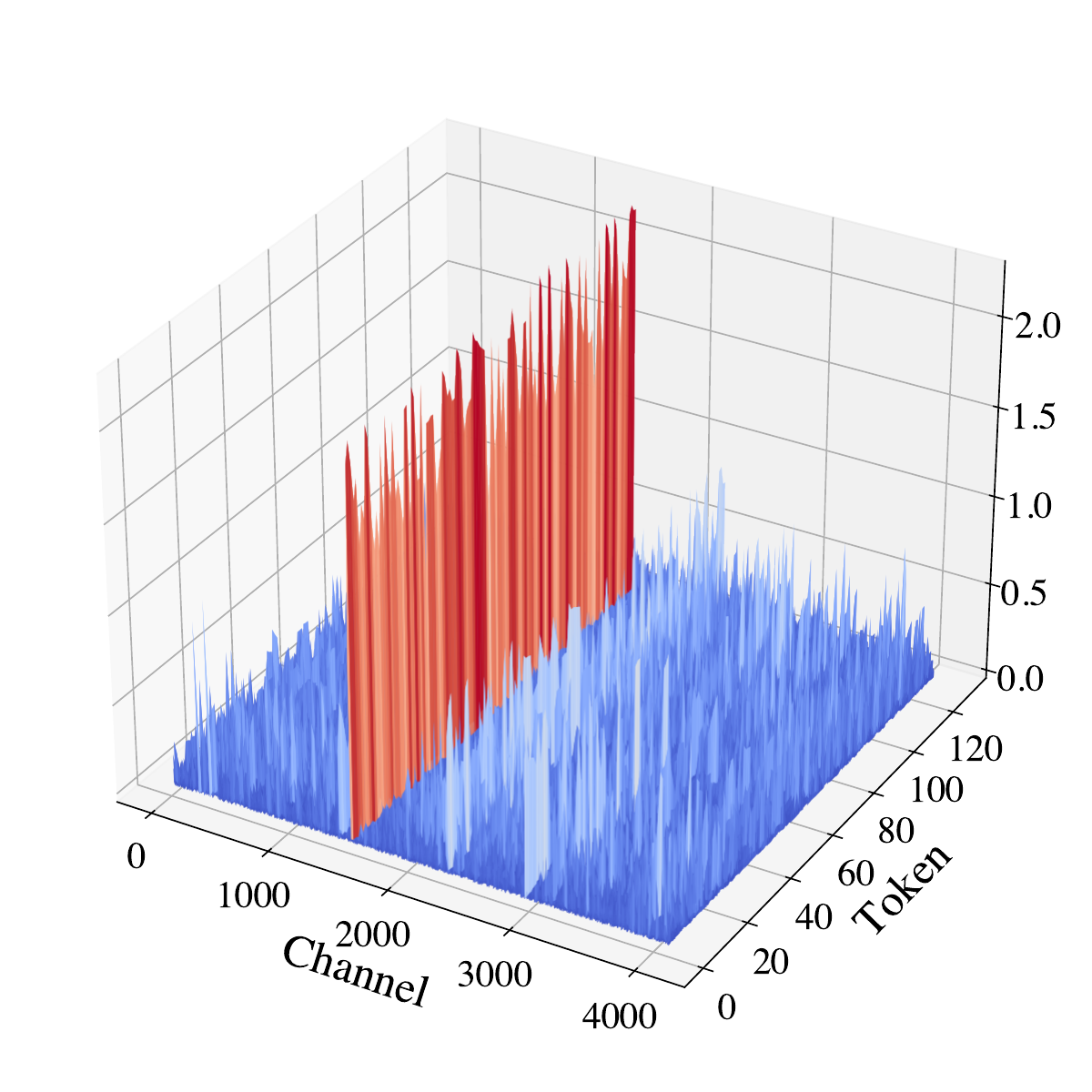}
        \label{fig:act_k_1_smooth}
    }
    \subfigure[Rotation]{
        \includegraphics[width=0.225\linewidth]{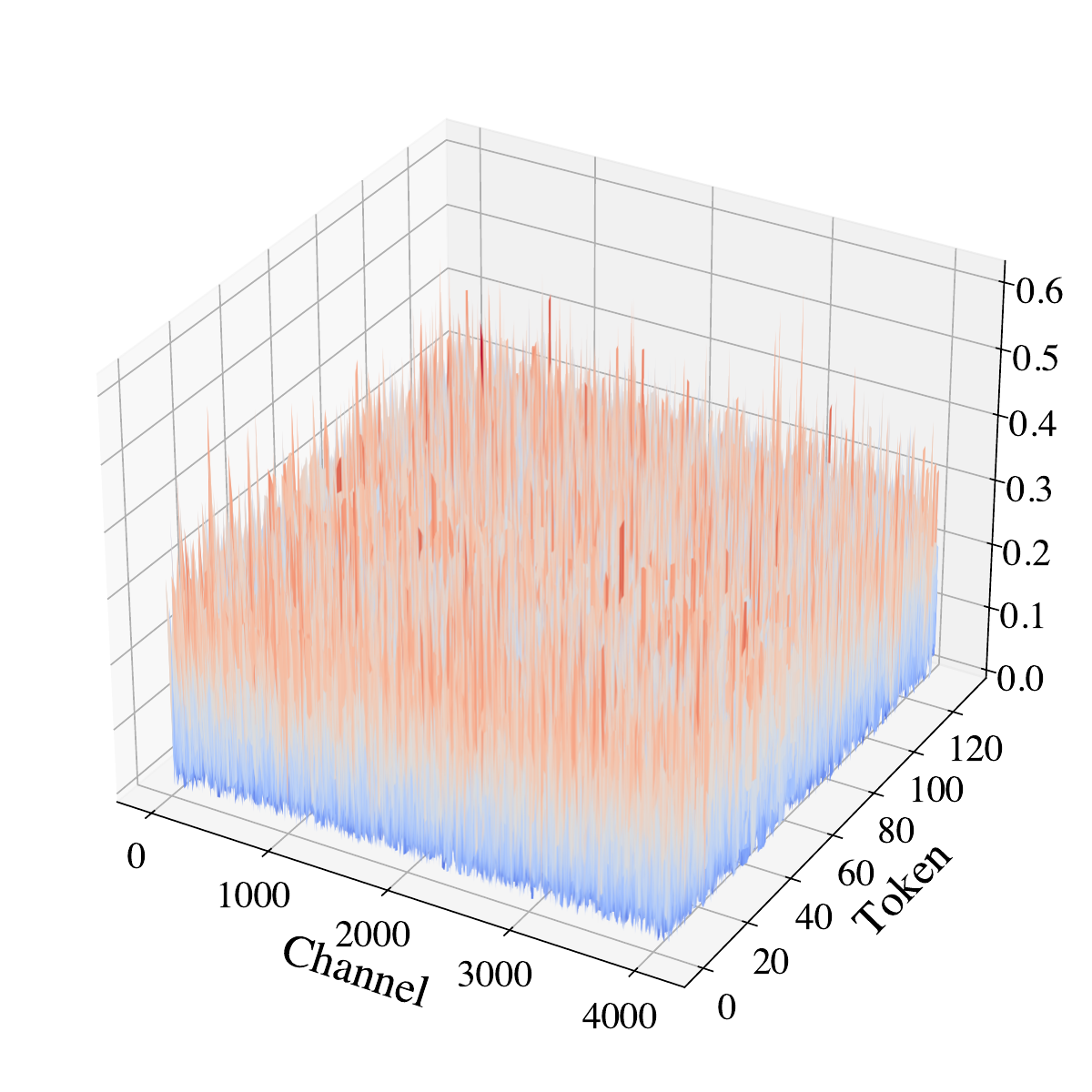}
        \label{fig:act_k_1_rot}
    }
    \subfigure[Smooth Rotation]{
        \includegraphics[width=0.225\linewidth]{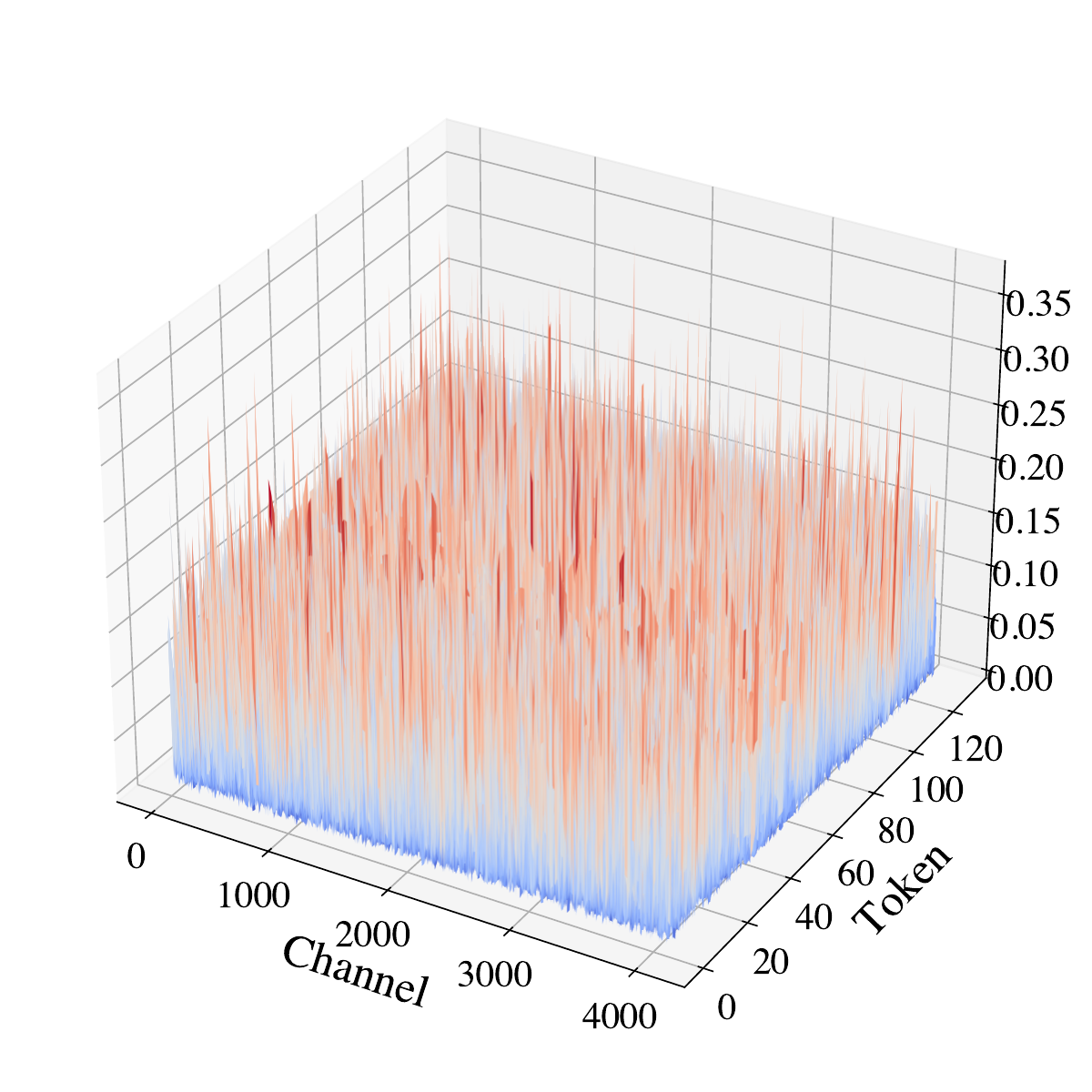}
        \label{fig:act_k_1_smooth_rot}
    }
    \caption{Input activation magnitudes measured at the second key projection layer (k\_proj 1) of LLaMA2-7B with different transformations applied. Note: query and value projection layers have the same input activation tensor.}
    \label{fig:act_k_1}
    \vspace{-0.15in}
\end{figure*}

\begin{figure*}[t]
    \centering
    \subfigure[Original]{
        \includegraphics[width=0.225\linewidth]{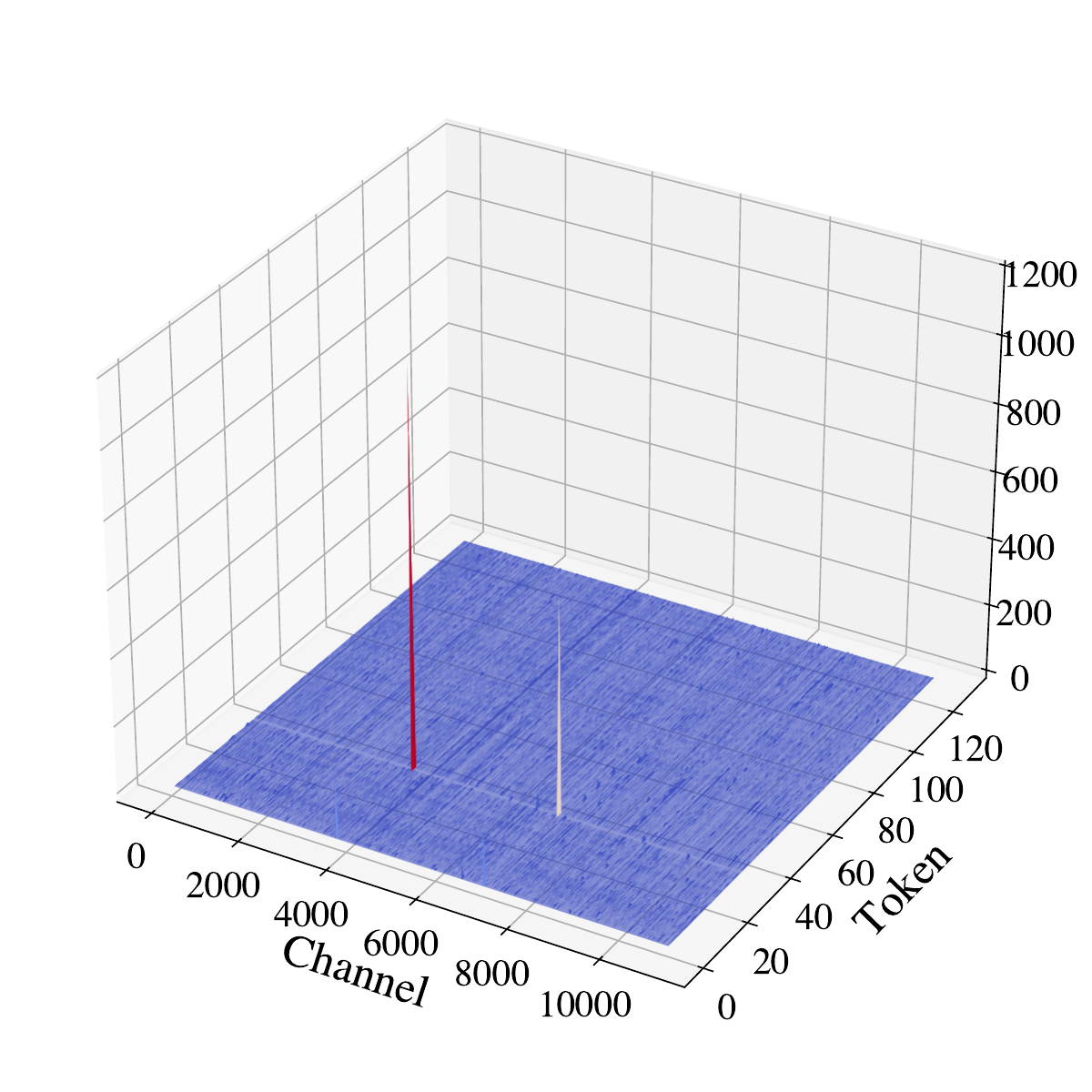}
        \label{fig:act_down_30_orig}
    }
    \subfigure[Smoothing]{
        \includegraphics[width=0.225\linewidth]{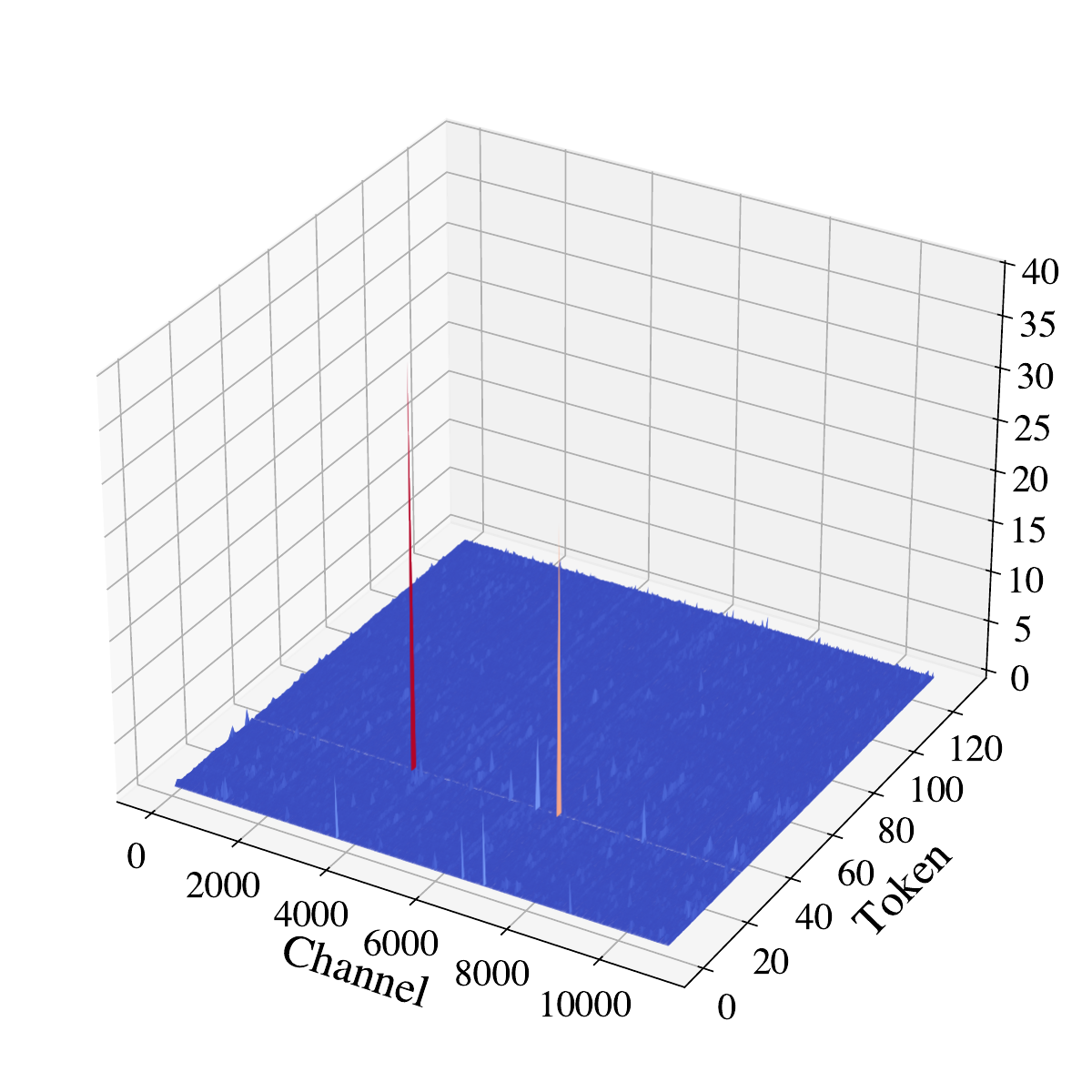}
        \label{fig:act_down_30_smooth}
    }
    \subfigure[Rotation]{
        \includegraphics[width=0.225\linewidth]{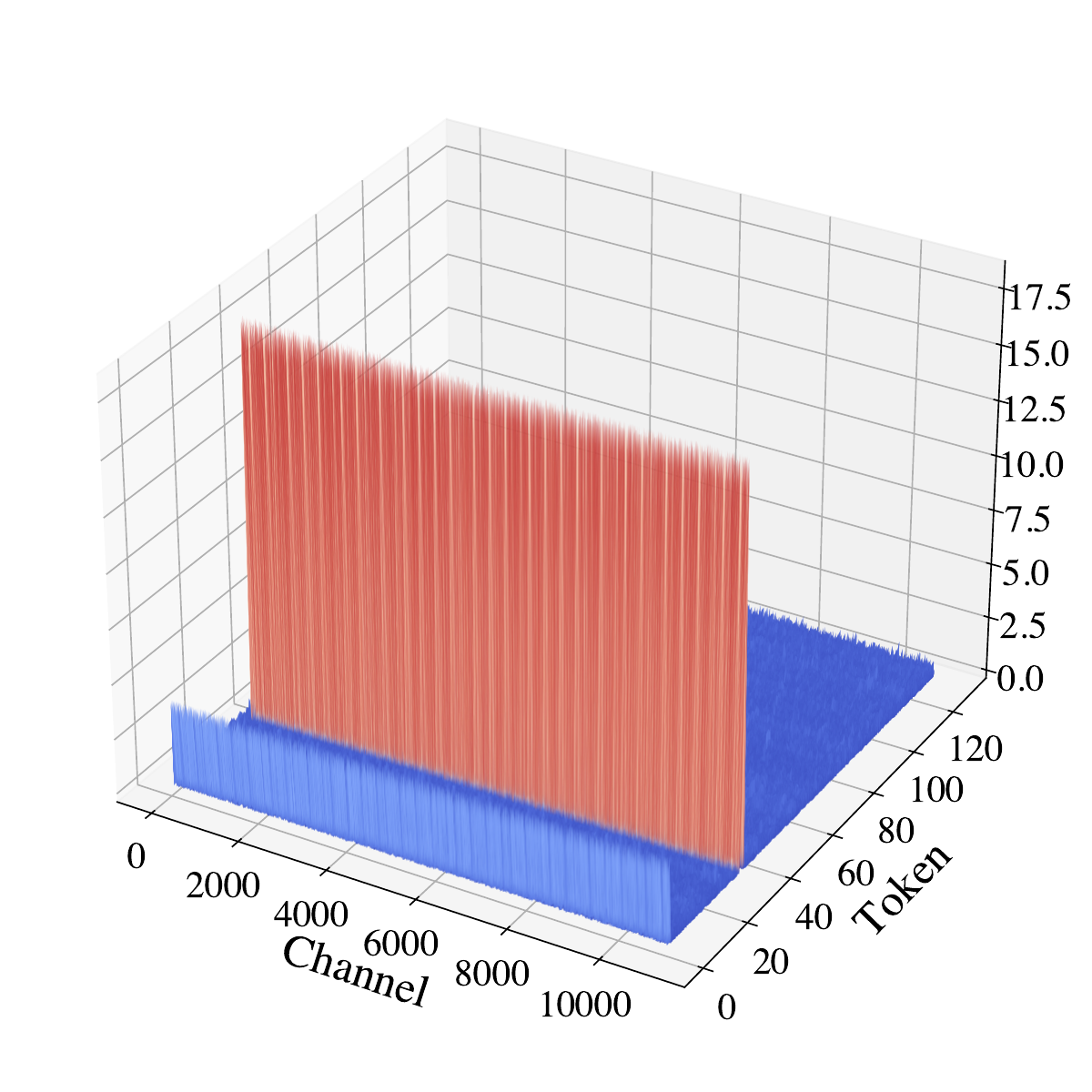}
        \label{fig:act_down_30_rot}
    }
    \subfigure[Smooth Rotation]{
        \includegraphics[width=0.225\linewidth]{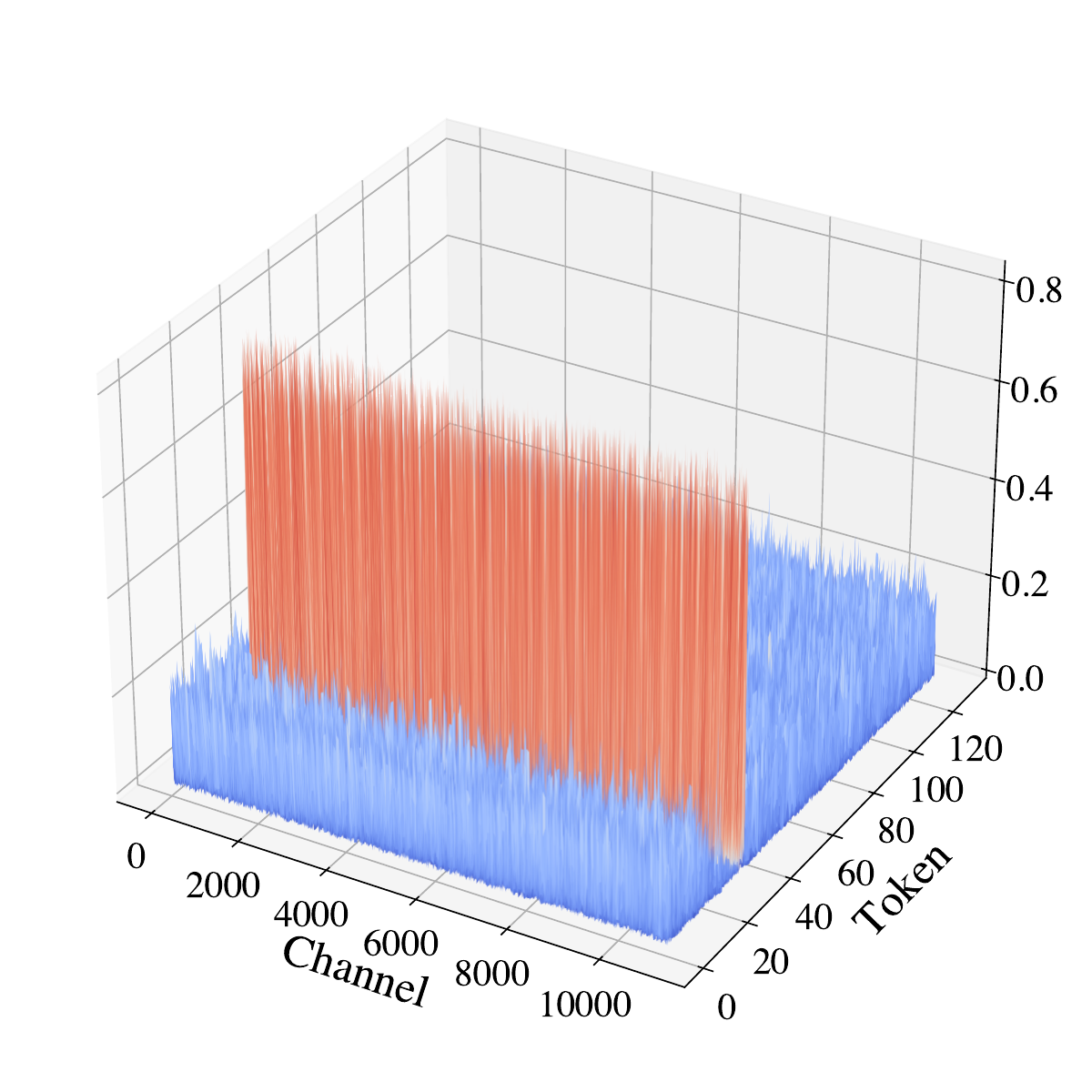}
        \label{fig:act_down_30_smooth_rot}
    }
    \caption{Input activation magnitudes measured at the second to last down projection layer (down\_proj 30) of LLaMA2-7B with different transformations applied.}
    \label{fig:act_down_30}
    \vspace{-0.15in}
\end{figure*}

We examine the effect of activation outliers on the quantization error of LLM layers and investigate the impact of different equivalent transformations. 

\subsection{Recording Activations}

We take a random sample from the WikiText-2~\cite{merityPointerSentinelMixture2017} dataset with a sequence length of 128 and propagate it through the LLaMA2-7B model~\cite{touvronLlama2Open2023}. We use the Hugging Face~\cite{wolfTransformersStateoftheArtNatural2020} implementation of the model with FP16 weights, registering PyTorch~\cite{paszkePyTorchImperativeStyle2019} hooks to capture input activations during forward propagation. The LLaMA2-7B model contains 32 decoder layers, each comprising an attention block and a Feed Forward Network (FFN) block. In every layer, we record the inputs to both blocks (k\_proj and gate\_proj) and the inputs to the modules producing the block outputs (o\_proj and down\_proj). Note that the key projection layer (k\_proj) shares its input with the value and query projections, similarly to how the gate projection (gate\_proj) in the FFN shares its input with the up projection. Consequently, we could have chosen those other modules for investigation without altering the observed activation distributions.

\subsection{Quantization}

We adopt standard uniform symmetric quantization using 4 bits for both activations and weights, with per-token and per-channel granularity, respectively. We do not apply any clipping to fully capture the effect of outliers. To maintain comparable quantization errors for activations and weights, we employ the RTN quantizer for both, which also helps reduce computational overhead.

\subsection{Channel-wise Scaling}

To obtain the channel-wise scaling factor $\mathbf{s}$, we use the formula introduced by SmoothQuant~\cite{xiaoSmoothQuantAccurateEfficient2023}:

\begin{equation} \label{eq:sq_s_definition}
s_j = \max\left(\lvert \mathbf{X}_j \rvert\right)^\alpha / \max\left(\lvert \mathbf{W}_j \rvert\right)^{1-\alpha},
\end{equation}

where $j$ denotes the channel index, and $\alpha$ is the migration strength controlling the extent to which quantization difficulty is shifted from the input activations to the weights. SmoothQuant identifies $\alpha=0.5$ as a sweet spot for various models, simplifying \eqref{eq:sq_s_definition} to $s_j = \sqrt{\max\left(\lvert \mathbf{X}_j \rvert\right) / \max\left( \mathbf{W}_j \right)}$. Since we do not calibrate the smoothing operation but apply it online (i.e., calculating the scaling factor based on the current data), we do not attempt to optimize alpha, even though this approach may result in suboptimal smoothing in some modules. Nevertheless, it helps prevent overfitting to the data sample, and we also benefit from the simplified formula when formulating the effects of transformations.

\subsection{Rotation}

To construct rotation matrices, we employ Hadamard matrices following QuIP\#~\cite{tsengQuIPEvenBetter2024} and QuaRot~\cite{ashkboosQuaRotOutlierFree4Bit2024}, but without additional randomization. A Hadamard matrix is an orthogonal square matrix of size $d=c_{in}$, having the following form:

\begin{equation} \label{eq:hadamard_matrix}
\mathbf{R} = \frac{1}{\sqrt{d}}[h_{i,j}]_{d\times d}, \quad \mathrm{where} \quad  h_{i,j} \in \{-1, +1\}.
\end{equation}

When $d=2^p$ (i.e., $d$ is a power of two), we use Sylvester construction, which recursively inflates the matrix with the Kronecker product using $R_{2^p} = R_2 \otimes R_{2^{p-1}}$, where $R_2$ is the initial 2×2 matrix. This approach is sufficient for almost all modules, where $c_{in} = 4096$, except for the down projection layers, where $c_{in} = 11008$. Although 11008 is not a power of two, it can be decomposed as $64 \times 172$, where $64 = 2^6$, and there exist known Hadamard matrices of size $172 \times 172$. Thus, we construct $R_{11008}$ as $R_{64} \otimes R_{172}$, adopting the technique from QuIP\#~\cite{tsengQuIPEvenBetter2024}. It is important to note that the columns of the Hadamard matrices constructed in this manner (with an infinitesimally small number of exceptions) have a mean of 0, meaning they contain an equal number of +1 and -1.

\section{Results and evaluation}

\begin{figure*}[t]
    \centering
    \subfigure[Layer-wise quantization error]{
        \includegraphics[width=0.305\linewidth]{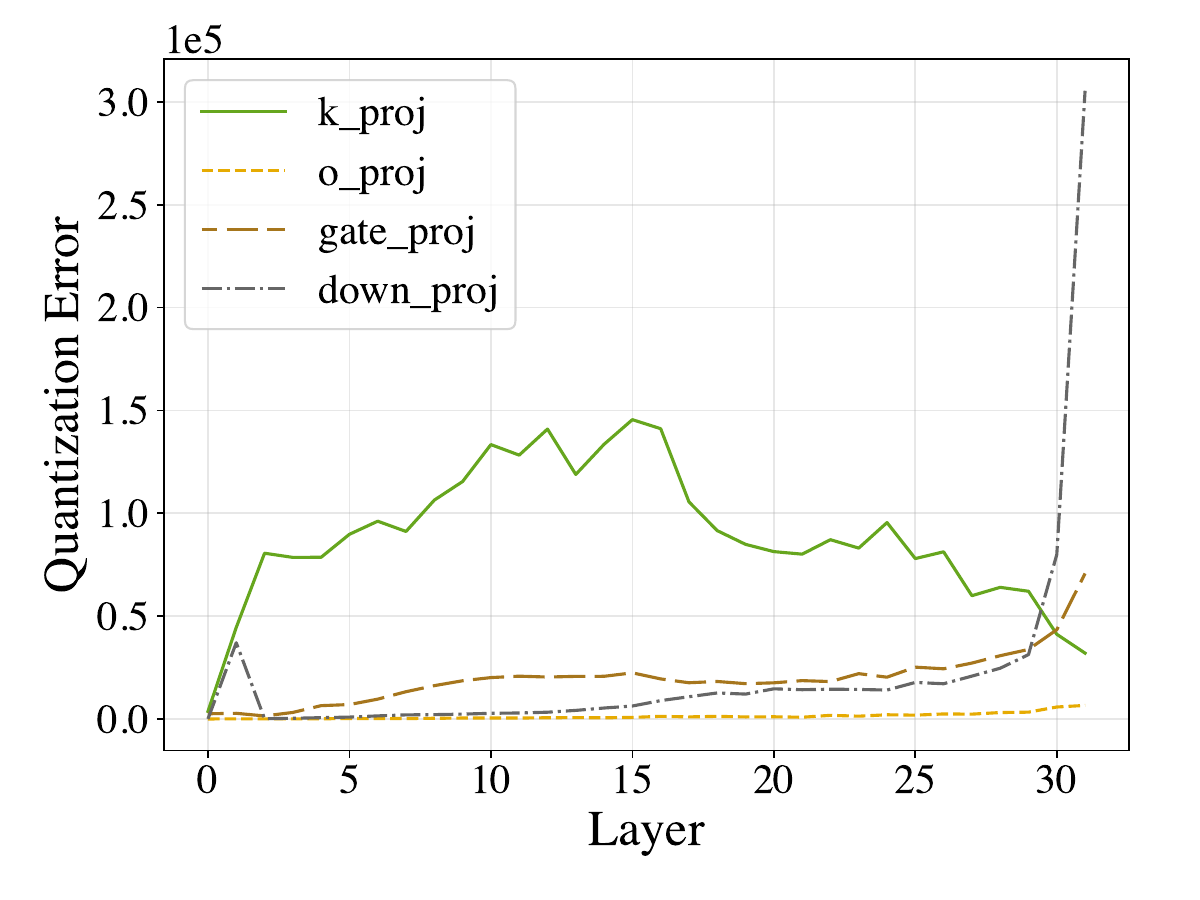}
        \label{fig:error_orig}
    }
    \subfigure[Activation quantization difficulty]{
        \includegraphics[width=0.305\linewidth]{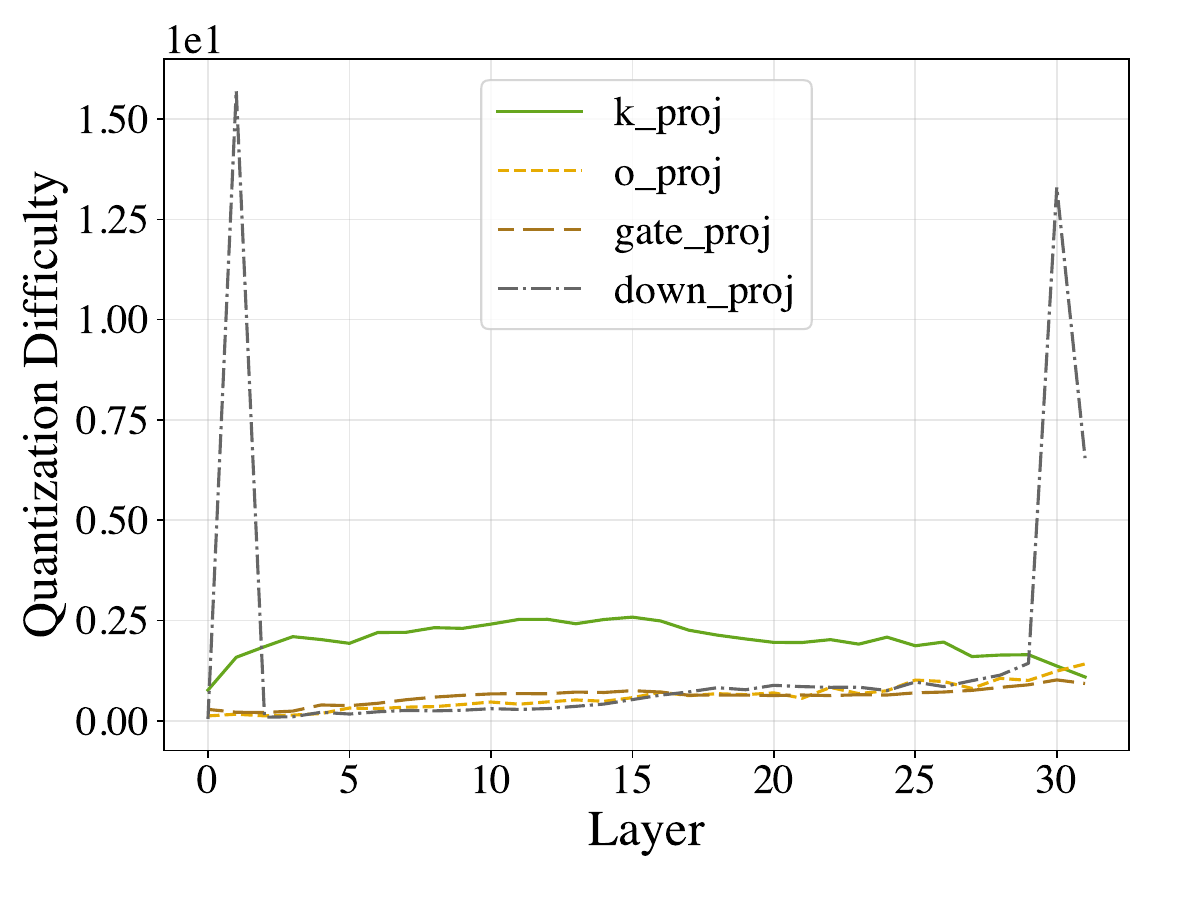}
        \label{fig:a_channel_magnitude_std_orig}
    }
    \subfigure[Weight quantization difficulty]{
        \includegraphics[width=0.305\linewidth]{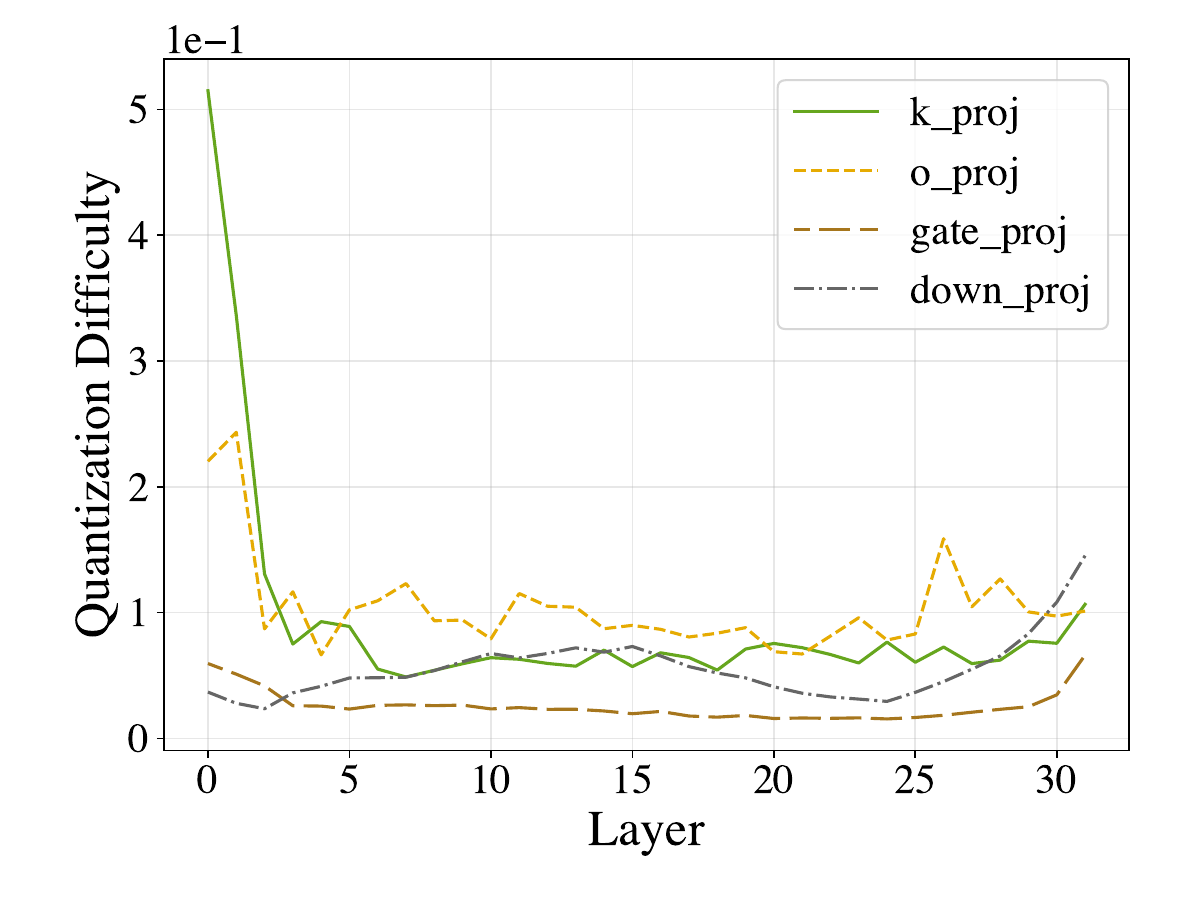}
        \label{fig:w_channel_magnitude_std_orig}
    }
    \caption{Layer-wise statistics measured at different modules of LLaMA2-7B.}
    \label{fig:stats_orig}
    \vspace{-0.15in}
\end{figure*}

\begin{figure*}[t]
    \centering
    \subfigure[Layer-wise quantization error]{
        \includegraphics[width=0.305\linewidth]{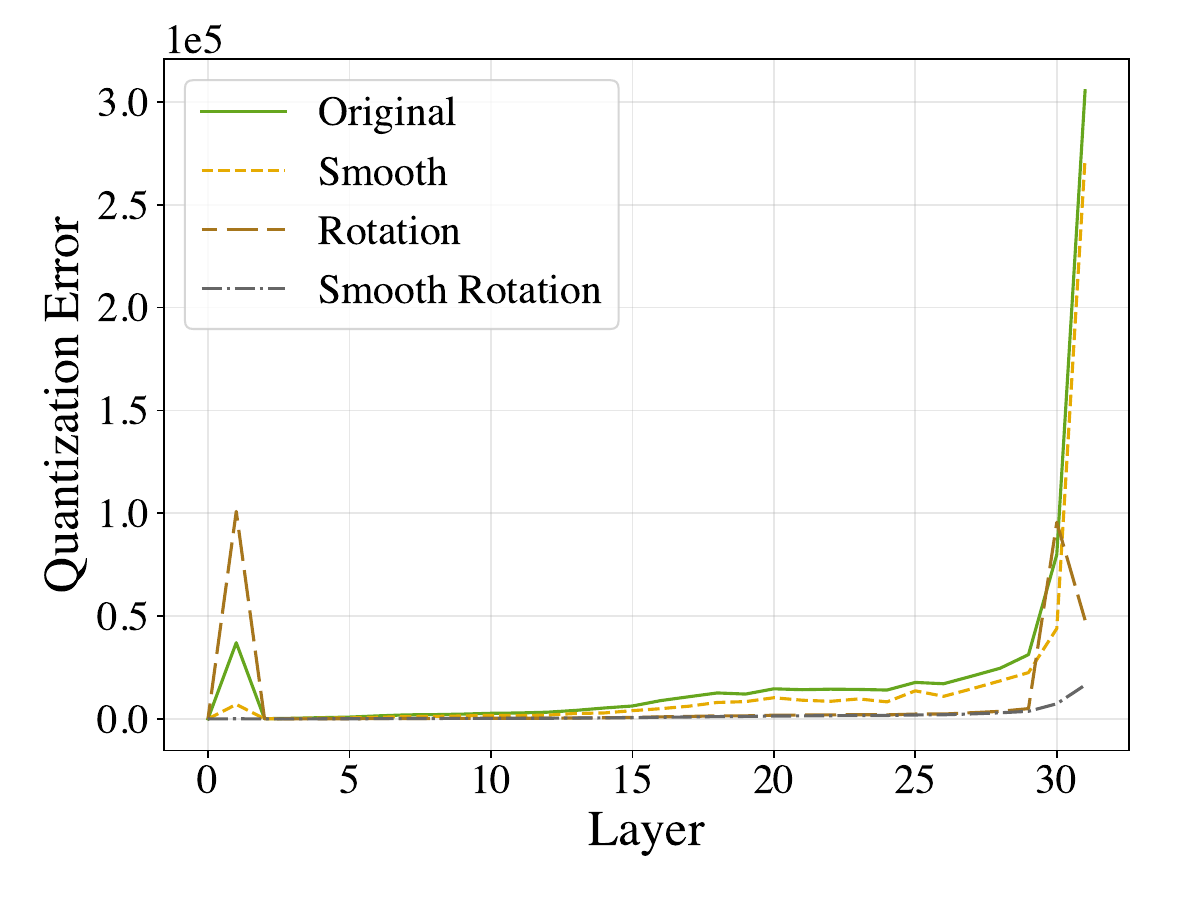}
        \label{fig:error_down_proj}
    }
    \subfigure[Activation quantization difficulty]{
        \includegraphics[width=0.305\linewidth]{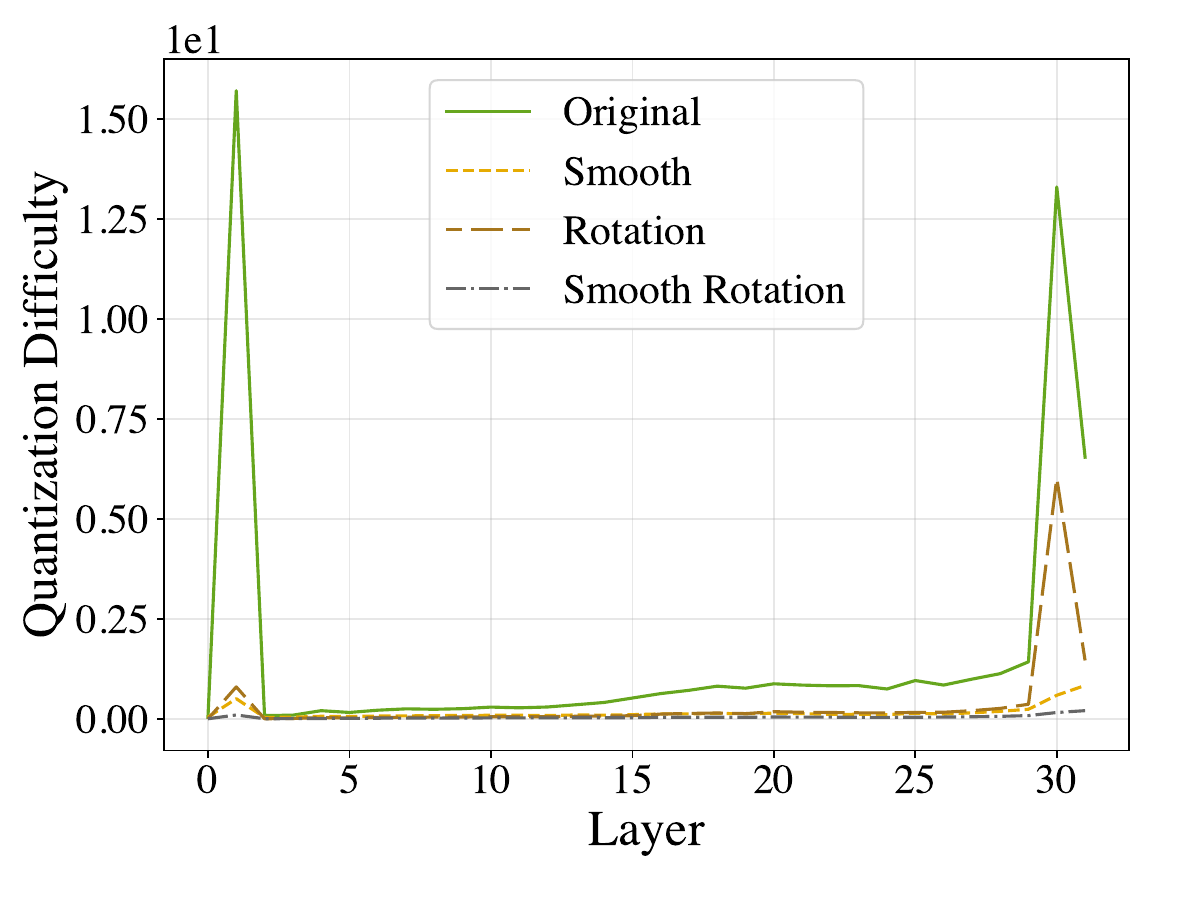}
        \label{fig:a_channel_magnitude_std_down_proj}
    }
    \subfigure[Weight quantization difficulty]{
        \includegraphics[width=0.305\linewidth]{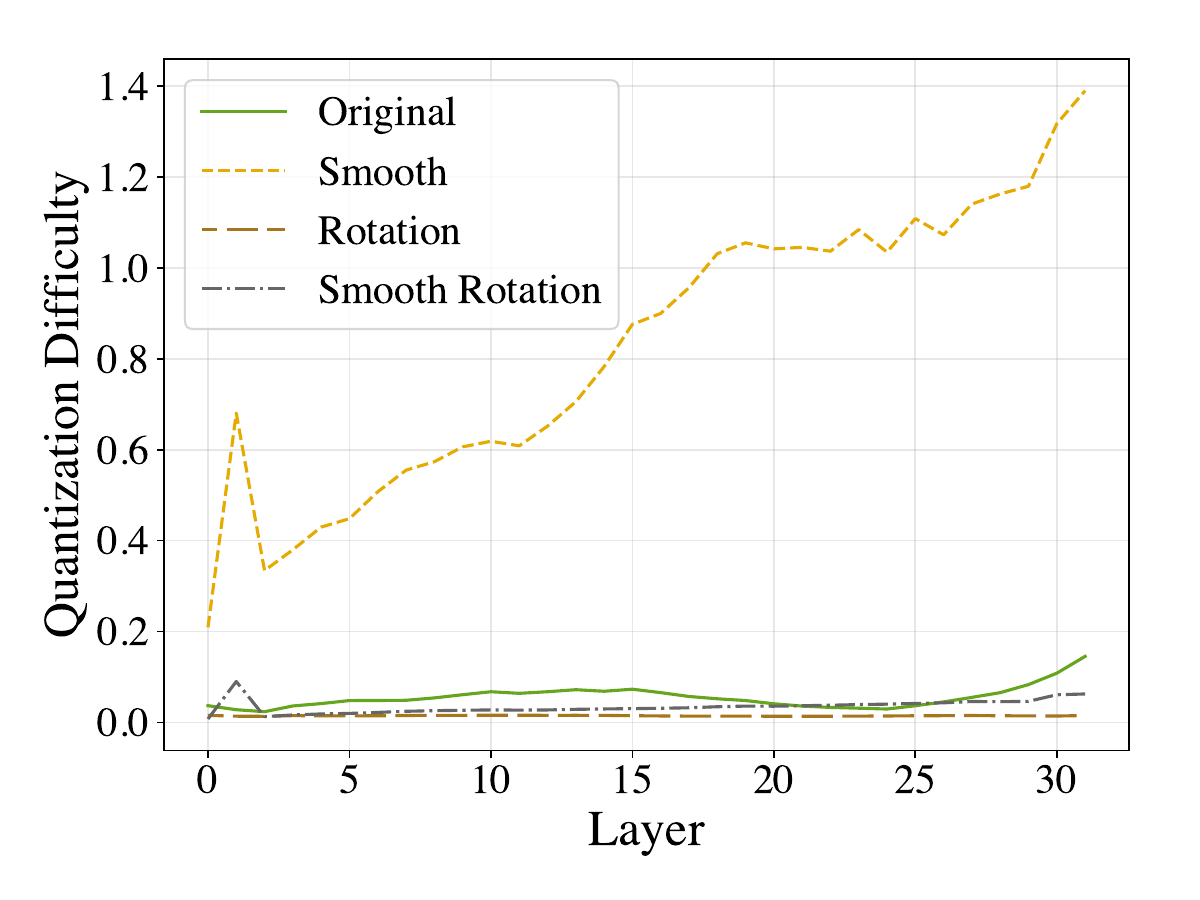}
        \label{fig:w_channel_magnitude_std_down_proj}
    }
    \caption{Layer-wise statistics measured at down projection layers of LLaMA2-7B with different transformations applied.}
    \label{fig:stats_down_proj}
    \vspace{-0.15in}
\end{figure*}


\subsection{Activation Distributions and Emergent Outliers}

The left-hand plots in \cref{fig:act_k_1,fig:act_down_30} depict the distribution of absolute activation values across various layers and modules. In line with previous studies, we identify two main types of activation outliers: systematic outliers~\cite{dettmersLLMint88bitMatrix2022, xiaoSmoothQuantAccurateEfficient2023}, which occur in a small number of channels across all tokens, and massive outliers~\cite{yangMitigatingQuantizationErrors2024, linDuQuantDistributingOutliers2024}, which are token-specific but exhibit considerably higher absolute values. Systematic outliers predominantly emerge in attention projection layers and the up/gate projections in the FFN block, whereas massive outliers appear almost exclusively in down projections, with notably large values (exceeding 1000) observed in the second and second-to-last decoder layers.

\subsection{Impact of Activation Outliers on Quantization Error}
\label{sec:orig_error}

\cref{fig:error_orig} presents the layer-wise quantization error in different modules. Although the quantization error of the attention key projection module increases until the midpoint of the model and then decreases, it grows in a nearly monotonic fashion for the other observed modules. Moreover, the FFN down projection layers with indices 1, 30, and 31 (i.e., the second and the last two layers) exhibit substantially higher out-of-trend errors. Building on the analytical methods introduced in FlatQuant~\cite{sunFlatQuantFlatnessMatters2024}, we define the quantization difficulty of a tensor as the standard deviation of its channel magnitudes. \cref{fig:a_channel_magnitude_std_orig,fig:w_channel_magnitude_std_orig} illustrate the resulting quantization difficulty for activations and weights, respectively. Upon omitting a small number of outliers (down\_proj 1/30/31 and gate\_proj 31), we observe a very strong correlation (over 0.97) between the layer-wise quantization error and the square of the activation’s quantization difficulty (i.e., the variance of channel magnitudes). This suggests that quantization error primarily depends on the underlying activation distributions in LLMs. While the omitted down projection layers also exhibit high activation quantization difficulty that correlates with elevated errors, their relationship is not entirely linear. In layers 1 and 30, this arises from the presence of massive outliers, which lead to significant quantization error in only a small portion of tokens, resulting in a lower overall error than the difficulty measure alone might predict. In contrast, the down projection module input in the last decoder layer contains large absolute value activations in multiple tokens, causing a notably high layer-wise quantization error despite having a comparatively lower channel magnitude-based quantization difficulty. Additionally, increased weight quantization difficulty further amplifies the overall error in this case, as it does in gate\_proj 31. Nonetheless, the quantization difficulty for weights is generally lower than it is for activations, suggesting that no substantial outliers occur in weight tensors.

\subsection{Effects of Channel-wise Scaling}
\label{sec:smoothing}

\cref{fig:stats_down_proj} displays the layer-wise quantization error and quantization difficulties of down projection layers with the tested transformations applied. Channel-wise scaling yields lower error than the original distributions; however, it generally falls short of rotation. Interestingly, in certain layers of the attention output projection and the FFN gate projection, the error exceeds that of the original distribution. We find that adjusting the migration difficulty ($\alpha$) can improve these results, as experiments indicate that the error can be kept below the original by choosing larger $\alpha$ values (approximately 0.7 for out\_proj and 0.65 for gate\_proj).

Turning to quantization difficulties, smoothing inherently produces flatter activation distributions than rotation in most cases. However, it significantly increases the quantization difficulty of the weight tensors, leading to less satisfactory results in 4-bit quantization. By substituting the formula \eqref{eq:eq_transformation} with \eqref{eq:sq_s_definition}, it becomes evident that the channel-wise maximum absolute values in both $\hat{\mathbf{X}}$ and $\hat{\mathbf{W}}$ can be expressed as $\sqrt{\max\left(\lvert \mathbf{X}_j \rvert\right) \cdot \max\left(\lvert \mathbf{W}_j \rvert\right)}$ (assuming $\alpha = 0.5$). This explains why their quantization difficulties are similar in both trend and magnitude.

\subsection{Effects of Rotation}

\begin{figure}[t]
    \centering
    \subfigure[Rotation]{
        \includegraphics[width=0.45\columnwidth]{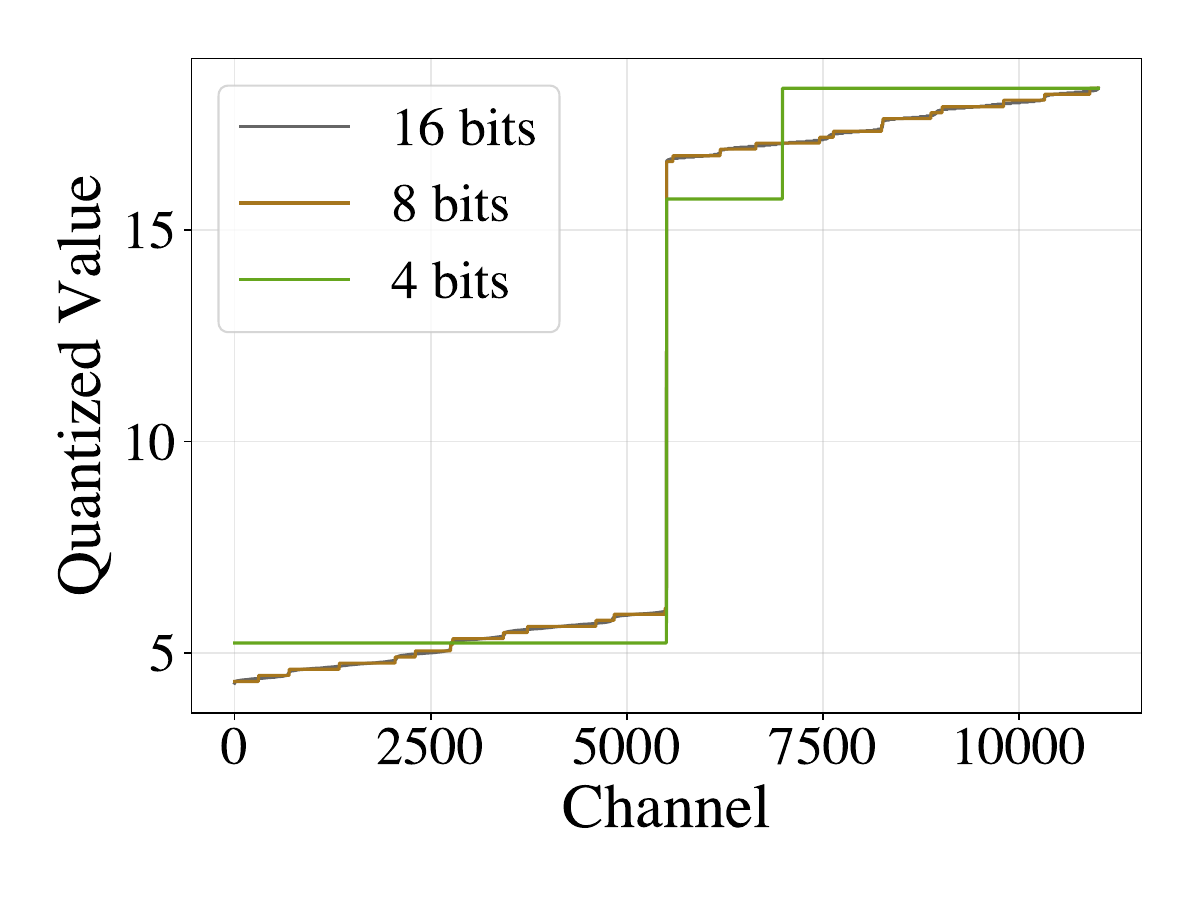}
        \label{fig:quant_bins_down_30_rot}
    }
    \subfigure[Smooth Rotation]{
        \includegraphics[width=0.45\columnwidth]{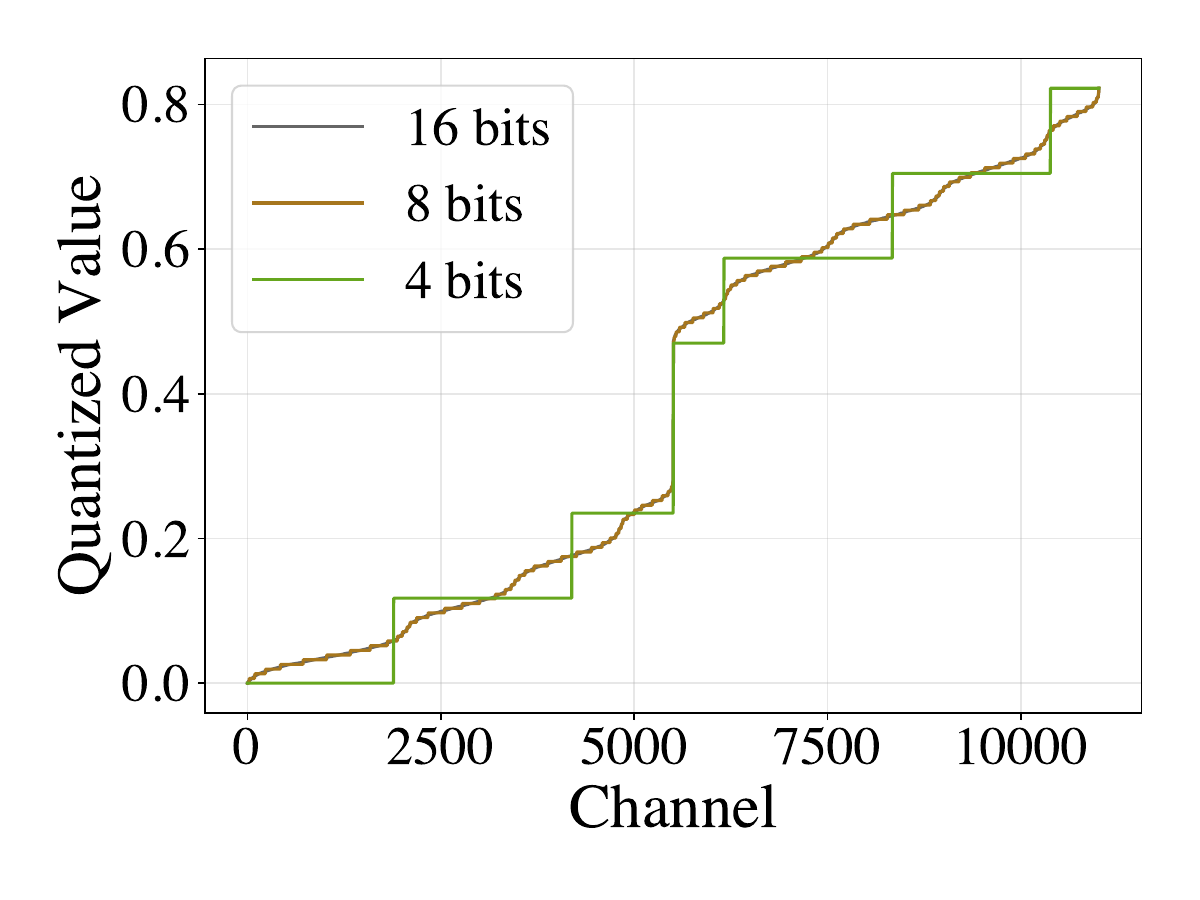}
        \label{fig:quant_bins_down_30_smooth_rot}
    }
    \vspace{-0.15in}
    \caption{Absolute value distribution and effective quantization bins in massive outlier tokens measured at the second to last down projection layer (down\_proj 30) of LLaMA2-7B with different transformations applied. Channels (horizontal axis) are sorted based on absolute value.}
    \label{fig:quant_bins_down_30}
    \vspace{-0.15in}
\end{figure}

Although rotated activations are less flat than their smoothed counterparts, they still exhibit substantially lower quantization difficulty than the original activations. Moreover, rotation not only avoids migrating difficulty to weights, but also redistributes weight values across input channels, resulting in weight tensors that are the most favorable for quantization, with even lower quantization difficulty than the original model. Consequently, rotation yields lower quantization error compared to smoothing.

However, as described in \cref{sec:orig_error}, the down\_proj 1 and 30 layers, featuring massive outliers, exhibit disproportionately high layerwise quantization error relative to their quantization difficulty. Counterintuitively, we find that the quantization error of these modules after rotation is even higher than without any transformation. Examination of the rotated distribution (\cref{fig:act_down_30_rot}) and the absolute values from the token with the largest magnitude (\cref{fig:quant_bins_down_30_rot}) in down\_proj 30 indicates that these values are concentrated around two distinct magnitudes (or one for the second layer). To understand the underlying cause, let us consider the token $\mathbf{t}$ containing massive outliers $o_j$ at dimensions $j \in O$, and $\epsilon \sim \mathcal{N}(0,\,\sigma^{2})$ at all other dimensions:

\begin{equation} \label{eq:massive_outlier_token}
    t_j \;=\; 
    \begin{cases}
        o_j, & \text{if } j \in O,\\[5pt]
        \epsilon, & \text{if } j \notin O,
    \end{cases}
    \quad \text{for } j = 1, \dots, d.
\end{equation}

We then formulate the rotated token $\mathbf{\hat{t}}$ using the Hadamard matrix R defined in \eqref{eq:hadamard_matrix}, taking advantage of the fact that R contains an equal number of positive and negative entries with the same absolute value in each column:

\begin{equation}
    \hat{t}_j = \mathbf{t} \cdot \mathbf{R}_j = \frac{\sum_{i \in O}h_{i,j}o_i}{\sqrt{d}} + \epsilon.
\end{equation}

This formula shows that the values in $\mathbf{\hat{t}}$ are clustered around $2^{|O|-1}$ distinct centroids (the number of possible $\pm$ sign combinations across the outlier dimensions), forming equally sized clusters with variance $\sigma^{2}$. This explains the distributions observed in the aforementioned figures and demonstrates that the absolute maximum value occurs in the dimensions $j, \text{where} \sum_{i \in O}|h_{i,j}| = |O|$, that is all outlier dimensions share the same sign in the rotation matrix, resulting in:

\begin{equation}
    \max(|\mathbf{\hat{t}}|) = \frac{\sum_{i \in O}|o_i|}{\sqrt{d}} + |\epsilon|.
\end{equation}

Therefore, the high quantization error arises because the dimensionality is insufficient to redistribute the outlier values in a way that lowers $ \max(|\mathbf{\hat{t}}|)$ enough to achieve a smaller quantization step size.

\subsection{Effects of Smoothing Before Rotation}

Recognizing that additional dimensions are needed for rotation to effectively flatten massive outliers, we first apply smoothing, thereby redistributing a portion of the outliers to the weights. We then apply rotation to these weights as well, effectively doubling the number of dimensions d through which the outlier values are spread and reducing the resulting maximum absolute values. Specifically, by smoothing and rotating the token as $\mathbf{\tilde{t}} = \mathbf{t} \cdot \mathrm{diag}(\mathbf{s})^{-1} \cdot \mathbf{R}$  with a migration strength of 0.5, the maximum absolute value changes to:

\begin{equation}
    \max(|\mathbf{\tilde{t}}|) \approx \sum_{i \in O}\sqrt{\frac{|o_i| \cdot \max(|\mathbf{W}_i|)}{d}}.
\end{equation}

\cref{fig:act_down_30_smooth_rot,fig:quant_bins_down_30_smooth_rot} illustrate how this process affects both the distribution and the quantization bins of outlier tokens. In terms of layer-wise quantization error, this hybrid method strikes a balance between smoothing and rotation, offering the lowest activation quantization difficulties while producing significantly flatter weight distributions than channel-wise scaling alone. It is particularly effective for massive outliers, as explained above. Although \cref{sec:smoothing} demonstrated that selecting an appropriate migration strength is crucial, our results show that even a fixed value of 0.5 achieves the lowest error among the transformations tested in most cases.

\section{Limitations and Future Work}

The primary limitation of our study is its constrained experimental scope, restricted to a single model and a small sample from one dataset. Nonetheless, these experiments still provide a reasonable degree of representativeness, given that major observations align with prior findings. Regarding the proposed method, it slightly increases the quantization difficulty of weights and depends on calibration in contrast to rotation alone. Consequently, based on our findings, we currently recommend Smooth Rotation only for down projection layers, where it effectively mitigates massive outliers, unless further evidence suggests otherwise. Finally, our evaluation only measured layer-wise error, without assessing end-to-end metrics such as perplexity.

To fully validate the positive impact of Smooth Rotation, future research should explore additional architectures (e.g., Mistral~\cite{jiangMistral7B2023}) and a range of model sizes. It would also be beneficial to examine the influence of different calibration sets on the results, as well as any potential perplexity improvements that the proposed technique might provide.
\section{Conclusion}

We investigate how activation outliers affect layer-wise quantization error and examine the influence of channel-wise scaling and rotation on outlier mitigation and overall error reduction. Specifically, we collect input activations from various modules in all layers of LLaMA2-7B, perform 4-bit integer quantization on both weights and activations, and measure quantization error layer by layer. In line with prior studies, we observe both systematic and massive outliers and their negative impact on activation quantization. We then apply smoothing and rotation transformations, finding that while rotation generally reduces quantization error more effectively, massive outliers can cause it to yield worse results than the unmodified model. To clarify these effects, we provide a mathematical formulation of how these transformations influence outlier magnitudes and propose a hybrid approach combining smoothing and rotation. Our results show that rotation mitigates smoothing’s weaknesses, while smoothing helps rotation to flatten distributions, yielding a balance between the two transformations when applied individually. This hybrid method significantly reduces quantization error in layers where massive outliers arise, consistently offering the lowest errors in other cases as well.

\section*{Acknowledgment}
The authors would like to thank the Doctoral School of Applied Informatics and Applied Mathematics of Obuda University for their valuable support.

\printbibliography

\end{document}